\documentclass[10pt,twocolumn,letterpaper]{article}

\usepackage[pagenumbers]{wacv} 

\usepackage{graphicx}  
\usepackage{adjustbox}
\usepackage{array}
\usepackage{pifont}
\usepackage{multirow}
\newcommand{\cmark}{\ding{51}}
\newcommand{\xmark}{\ding{55}}
\usepackage{tabularx}
\usepackage{siunitx} 

\definecolor{wacvblue}{rgb}{0.21,0.49,0.74}
\usepackage[pagebackref,breaklinks,colorlinks,allcolors=wacvblue]{hyperref}

\def\wacvPaperID{1422} 
\def\confName{WACV}
\def\confYear{2026}

\title{LangPose: Language-Aligned Motion for Robust 3D Human Pose Estimation}

\author{Longyun Liao\\
McMaster University\\
{\tt\small liaol13@mcmaster.ca}
\and
Rong Zheng\\
McMaster University\\
{\tt\small rzheng@mcmaster.ca}
}

\begin{document}
\maketitle
\begin{abstract}
2D-to-3D human pose lifting is an ill-posed problem due to depth ambiguity and occlusion. Existing methods relying on spatial and temporal consistency alone are insufficient to resolve these problems especially in the presence of significant occlusions or high dynamic actions. Semantic information, however, offers a complementary signal that can help disambiguate such cases. To this end, we propose LangPose, a framework that leverages action knowledge by aligning motion embeddings with text embeddings of fine-grained action labels. LangPose operates in two stages: pretraining and fine-tuning. In the pretraining stage, the model simultaneously learns to recognize actions and reconstruct 3D poses from masked and noisy 2D poses. During the fine-tuning stage, the model is further refined using real-world 3D human pose estimation datasets without action labels. Additionally, our framework incorporates masked body parts and masked time windows in motion modeling, encouraging the model to leverage semantic information when spatial and temporal consistency is unreliable. Experiments demonstrate the effectiveness of LangPose, achieving SOTA level performance in 3D pose estimation on public datasets, including Human3.6M and MPI-INF-3DHP. Specifically, LangPose achieves an MPJPE of 36.7mm on Human3.6M with detected 2D poses as input and 15.5mm on MPI-INF-3DHP with ground-truth 2D poses as input.
\end{abstract}
    
\section{Introduction}
\label{sec:intro}

3D human pose estimation (HPE) is crucial for better understanding human motion and serves as a cornerstone for numerous downstream tasks such as human action recognition, human-computer interaction, and virtual reality. 2D-to-3D pose lifting, which estimates 3D human poses from 2D keypoints extracted from a single image or video frame, is a mainstream approach to 3D human pose estimation due to the maturity and computational efficiency of modern 2D pose detectors. However, it is an ill-posed problem as depth ambiguity and occlusion often result in multiple possible 3D poses corresponding to a single 2D pose.

To cope with the inherent ambiguity, prior work has relied on spatial and temporal cues to enforce consistency in 3D pose predictions. Various architectures, including TCN-based~\cite{3dhpe_tcn, 3dhpe_st_explicit_occlusion_training}, GCN-based~\cite{2d-3d-gnn-hpe, ci_3dhpe_gcn, motion_guided_3dhpe_gcn}, and transformer-based models~\cite{mixste, motionbert2022, peng2024ktpformer}, have been designed to capture this information. Pretraining strategies such as random frame masking, joint masking, and noise addition on large-scale dataset have been utilized to enhance the learning of spatial and temporal cues~\cite{motionbert2022}. 

However, spatial and temporal consistency diminishes in the presence of significant occlusions or high dynamic actions such as those found in the real world (\eg, in sports). Semantic information refers to high-level contextual knowledge about human actions, intentions, or interactions. Such information, often available in fine-grained textual descriptions of human activities, can inform or constrain pose estimation and help disambiguate pose configurations, particularly in cases of occlusion and depth ambiguity. Semantic information can serve as a complementary cue. For example, when only the upper body is visible, semantic priors associated with the action ``sitting down” indicate that the hips are flexed and the knees are bent to roughly $90^o$, allowing the model to plausibly reconstruct the occluded lower‑limb joints. Similarly, in highly dynamic scenes, where temporal cues become less reliable, semantic priors linked to the ongoing action, for example, the expected limb trajectories and contact events in ``dribbling a ball", can guide the estimation of a full pose sequence even if it is partially occluded over time. 

One possible approach to incorporating action cues into 3D human pose estimation is through multi-task learning (MTL), where pose estimation and action recognition are jointly optimized. However, MTL typically represents action categories with one‑hot vectors, thereby discarding the rich semantic relationships conveyed in fine‑grained textual descriptions. Furthermore, such models do not generalize well to pose sequences from unseen action classes. Therefore, we argue for {\it explicitly} aligning the latency feature space of pose sequences with action descriptions and propose LangPose, a framework that learns a shared pose–language embedding via cross-modal contrastive training.  
Datasets designed for human motion generation, such as BABEL~\cite{BABEL:CVPR:2021} and KIT~\cite{kit}, offer richer and more fine-grained action descriptions. However, they introduce new challenges: (1) severe class imbalance—word like `walk' and  `transition' occur with high frequency, while many other descriptions are underrepresented in BABEL~\cite{BABEL:CVPR:2021}; and (2) the non-uniqueness of textual action descriptions, as the same motion can correspond to multiple valid descriptions, especially in the presence of concurrent or compound actions, and vice versa.


To address the above challenges, LangPose leverages a two-stage training process comprising a pretraining phase followed by fine-tuning. In the pretraining stage, the model is trained on a synthetic dataset~\cite{BABEL:CVPR:2021} containing motion sequences paired with fine-grained action labels, jointly optimizing for semantic alignment and 3D pose reconstruction. The semantic knowledge acquired during pretraining is generalizable across different 3D pose estimation datasets, in contrast to previous methods~\cite{2D3D_multi_task, multi_task_realtime, actionPrompt} that require retraining an action classifier for each target dataset. To tackle the issues of imbalanced data distribution in the pretraining dataset and the inherent uncertainty in human action recognition tasks, we replace the cross-entropy loss in infoNCE with a combination of KL-divergence and focal loss~\cite{focal_loss} to supervise multi-modal alignment. Additionally, inspired by MotionBERT~\cite{motionbert2022}, we implement varying-size temporal window masking and body part-level masking during pretraining. These strategies further encourage the model to reconstruct poses given semantic information rather than relying solely on spatial and temporal consistency. By the end of the pretraining stage, the model is equipped to recognize actions. In the fine-tuning stage, the model is further refined using real-world 3D human pose estimation datasets without action descriptions.

In summary, the contributions of this work are three folds:

\begin{itemize}  
    \item To the best of our knowledge, this is the first work to perform multi-modal pretraining that directly aligns the text embedding of descriptive action labels with skeleton motion sequences for a 3D human pose estimation task.
    \item We propose a new pre-training and fine-tuning strategy for motion representation learning. By integrating multi-modal representation learning of motion with descriptive action labels and 3D human pose estimation, the model is trained to embed generalizable action cues into 3D pose estimation during the pretraining stage.
    \item Experiments show that LangPose achieves competitive results on public benchmarks, surpassing most methods on MPI-INF-3DHP and attaining SOTA-level performance on Human3.6M, with notable gains on challenging actions and lower inference cost than diffusion-based models.
    
\end{itemize}
\section{Related work}
\label{sec:relatedwork}

\subsection{3D Human pose estimation}

3D human pose estimation can be categorized into two main approaches: the first involves estimating 3D human poses directly from images~\cite{ 3dHpeCNN, integral_hpe, 3dHPE_multiperson_single_rgb_image}, and the second involves lifting 2D poses to 3D poses~\cite{martinez_2017_3dbaseline, attention_temporal_contexts_3d_hpe, mixste,   peng2024ktpformer}. This work falls into the second category. Within this category, various architectures have been proposed to capture spatial and temporal information in motion dynamics, including TCN-based~\cite{3dhpe_tcn, 3dhpe_st_explicit_occlusion_training}, GCN-based~\cite{2d-3d-gnn-hpe, ci_3dhpe_gcn, motion_guided_3dhpe_gcn}, and transformer-based~\cite{ mixste, motionbert2022, peng2024ktpformer, tcpformer} models.

Since action labels provide comprehensive summaries of motion, we adopt spatial and temporal transformers~\cite{motionbert2022} as our backbone for the pose encoder, which effectively learns and fuses motion representations across both temporal and spatial dimensions.

\subsection{Skeleton-based human action recognition}

Another task related to understanding human motion is human action recognition. Unlike 3D human pose estimation, which focuses on precise pose reconstruction, human action recognition aims to comprehend human motion at a higher level. Despite the different objectives, both tasks require an understanding of the spatial relationships between the movements of different body joints and the temporal dynamics within motion. Consequently, recent works have designed various architectures to capture the spatio-temporal information of motion dynamics for human action recognition including transformer-based~\cite{star-transformer-HAR, motionbert2022, do2024skateformer}, GCN-based~\cite{ st-gcn,  channel-gcn-har, infoGCN, zhou2024blockgcn}, LSTM-based~\cite{lstm_har2016, lstm-har}, and CNN-based~\cite{view-cnn-har, topo-cnn-har}.
 
\begin{figure*}
    \centering
    \includegraphics[width=0.65\linewidth]{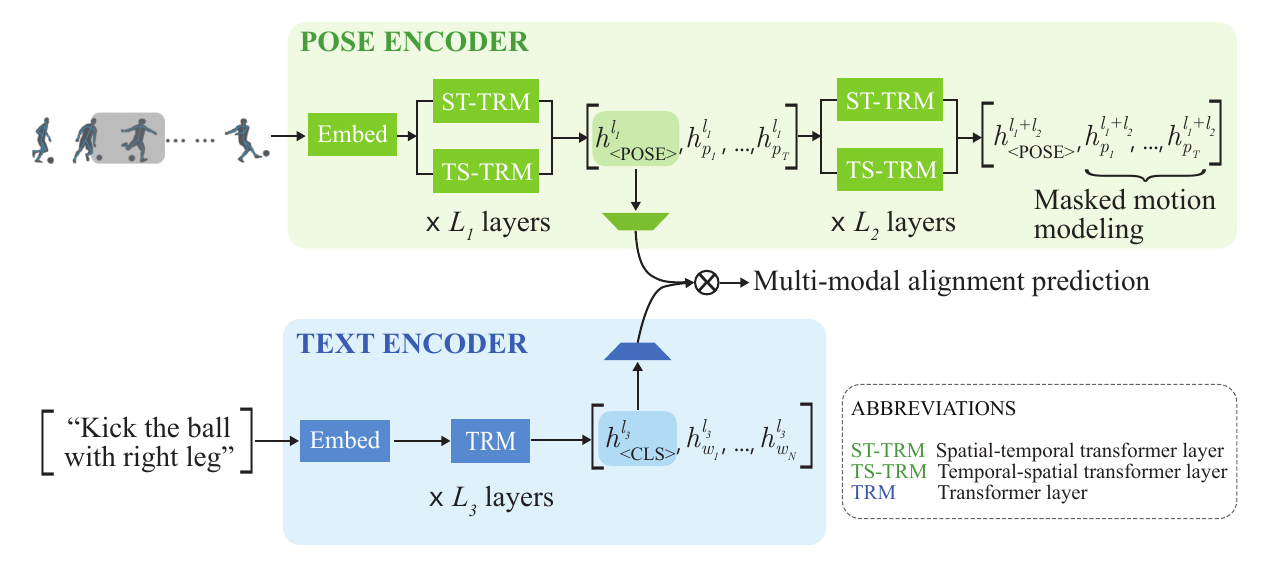}
    \caption{The overall architecture of LangPose.}
    \label{fig:enter-label}
\end{figure*} 

\subsection{Multi-task learning}

We use multi-task learning when the knowledge gained from one task can benefit other tasks~\cite{multi_task_survey}. This is particularly relevant for human pose estimation and human action recognition. The work in~\cite{2D3D_multi_task} was the first to highlight the interconnections between these tasks and attempted to estimate 2D/3D human poses alongside human action recognition. In their approach, both tasks share a common multi-task CNN, with human action recognition having an additional head to classify actions based on estimated poses and visual features. However, their method does not fully utilize the rich semantic information of actions, as the action classes are encoded as one-hot vectors. Similarly,~\cite{multi_task_realtime} aims to estimate 2D and 3D human poses and corresponding actions in real-time, but also encodes action classes as one-hot vectors.

\subsection{Multi-modal representation learning}
\label{section: related_work_multimodal}

Multi-modal representation learning methods, such as CLIP~\cite{CLIP} and ALIGN~\cite{pmlr_scale_up_vl}, have proven highly effective for tasks like image captioning, text-image retrieval, and zero-shot image classification. These methods work by aligning text embeddings from a text encoder with feature embeddings from an image encoder. This alignment allows the image encoder to capture and utilize rich semantic information from images, significantly enhancing its performance on downstream tasks.

Pioneer works point out that similar techniques can be applied broadly across the motion–language spectrum, including text-to-motion generation~\cite{motionclip, petrovich22temos, petrovich21actor}, motion captioning~\cite{Fang2025HuMoCon}, text–motion retrieval~\cite{petrovich23tmr, t2m_retrieval}, and text-aided discriminative tasks such as action recognition~\cite{ActionCLIP_skeleton} and pose estimation~\cite{actionPrompt, poseembroider, feng2024chatpose}. Building on these task-specific advances, a recent trend is to pursue general-purpose motion–language models that unify generation, captioning, retrieval, and discriminative perception within a single pretrained backbone (e.g., shared encoders with lightweight task heads)~\cite{li2025lamp, jiang2024motiongpt, wu2025motionagent}. In parallel, substantial effort has gone into curating large-scale motion–language corpora and whole-body resources to enable such unified modeling~\cite{Guo_2022_humanml3d,kit, BABEL:CVPR:2021, lin2023motionx, zhang2025motionx++}.


Among the studies above, the most relevant to our setting are two lines of work that reconstruct 3D motion from different inputs: multi-modal 3D human pose estimation from 2D to 3D, and text-conditioned motion generation from text to 3D. Our approach differs in two ways. First, prior multi-modal 3D HPE with textual cues usually centers on images rather than videos~\cite{poseembroider, feng2024chatpose}, so the semantics of motion dynamics remain underexplored for the 3D HPE task. Second, our method complements text-conditioned generation without performing generation: rather than synthesizing motion from language, we leverage motion semantics captured by text–motion alignment as auxiliary supervision to regularize 2D-to-3D pose estimation, preserving camera-aware, metric accuracy while adding semantic guidance, especially under occlusion. In this way, we complete the picture—from learning semantics to using semantics for reconstruction.
 

The most closely related approach that aligns motion and text for human pose estimation is ActionPrompt~\cite{actionPrompt}. Their method consists of two main components: an action-related text-prompt block and an action-specific pose-prompt block. The former enables the pose encoder to predict action labels from pose sequences using supervision from human pose estimation datasets~\cite{h36m_pami, humaneva}, while the latter refines the predicted poses based on the identified action. However, the semantic representations learned by the action-related text-prompt block are not transferable across datasets and must be retrained for each new domain. Moreover, since action labels in pose estimation datasets are typically coarse, this approach relies heavily on the action-specific pose prompts to recover fine-grained semantic details. This two-stage pipeline further assumes accurate classification of motion into discrete action labels, which is prone to failure in the presence of severe noise.
In contrast, LangPose performs end-to-end alignment between motion and semantic information while reconstructing 3D pose sequences. The learned semantics are generalizable across datasets and eliminate the need for storing action-specific pose prompts during inference.

\section{Methodology}

\subsection{Network architecture}
The overall objective of LangPose is to enable 3D human pose estimation and text-motion alignment. The network consists of two encoders to extract features from text and pose data. Embedding pooling layers are then applied to generate global representations of the text and pose. These representations are used for text-motion alignment, and a regression head subsequently projects the pose features to estimate the 3D pose. The network architecture is illustrated in~\cref{fig:enter-label}.

\noindent
\textbf{Text and pose encoders. } The LangPose network consists of two parallel BERT-style models~\cite{devlin2018bert} operating over pose and text domains. The text stream utilizes three layers of transformer blocks, following the architecture of the original BERT~\cite{devlin2018bert}. The pose stream employs five layers of Spatial-Temporal transformer blocks, following MotionBERT~\cite{motionbert2022}.

Given a sequence of 2D pose skeletons $\mathbf{x} \in \mathbb{R}^{T \times J \times C_{in}} $, we first project them into a higher dimensional space through one MLP layer, resulting in features $\{p_1, p_2, ..., p_T\}$ where $p_i \in \mathbb{R}^{J \times C_{f}}$, $C_{in}$ is the input dimension of 2D pose sequences, $C_{f}$ is the feature dimension of pose embeddings. We then concatenate a learnable pose class token to these features, forming the sequence $\{p_0, p_1, p_2, ..., p_T\}$ where $p_0$ is the class token \texttt{<POSE>}.

For the text input, we have $\{w_0, w_1, w_2, ..., w_N\}$, where $w_0$ is \texttt{<CLS>} and $w_N$ is \texttt{<SEP>}, namely tokens for classification and sentence separation. After processing the text inputs through the text encoder, our model generates encoded text features $\{h_{w_0}^{l_3}, h_{w_1}^{l_3}, h_{w_2}^{l_3}, ..., h_{w_N}^{l_3}\}$. Similarly, after processing the pose sequences through the pose encoder, our model produces encoded pose features $\{h_{p_0}^{l_1 + l_2}, h_{p_1}^{l_1 + l_2}, h_{p_2}^{l_1 + l_2}, ..., h_{p_T}^{l_1 + l_2}\}$.

Next, we introduce the structures to derive the global text representation and global pose representation, as follows.

\noindent
\textbf{Text embedding and pose embedding pooling layers for late fusion. } The purpose of these layers is to project the learned features of the pose class token \texttt{<POSE>} and the text class token \texttt{<CLS>} to the same feature space. These layers consist of multiple MLPs (Multi-Layer Perceptrons). 

Instead of average pooling all pose features, we use the pose feature in the pose class token, as it represents a weighted summation of all pose features through attention. Since pose features include an additional joint dimension, we perform a learnable weighted summation over the joint dimension of the pose class token \texttt{<POSE>}.

We extract the global pose representation from the middle layer of the pose encoder $h_{{<\texttt{POSE}>}}^{l_1}$. Shallow layers are chosen because they capture more action-specific information, which can then be propagated to deeper layers for 3D human pose estimation. After this stage, we denote the text global representation as $h_W$ and the pose global representation as $h_P$ for simplicity. These global representations are used in the task of multi-modal alignment prediction.

\noindent
\textbf{Pose estimation regression head. } This structure is designed to project pose features $\{h_{p_1}^{l_1 + l_2}, h_{p_2}^{l_1 + l_2}, ..., h_{p_T}^{l_1 + l_2}\}$ to predict 3D pose sequence $\mathbf{X} \in \mathbb{R}^{T \times J \times 3}$. 

\subsection{Pretraining tasks}

We introduce two new pretraining strategies in addition to random joint and frame-level masked motion modeling: to promote semantic reasoning under weak spatial and temporal consistency and to improve the ability to classify actions using partial body poses.

\begin{figure}
    \centering
    \includegraphics[width=0.6\linewidth]{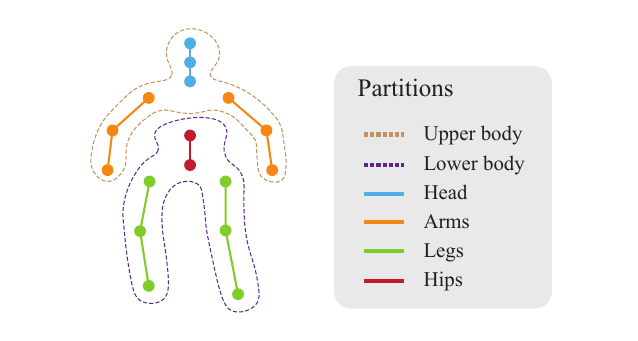}
    \caption{The human body is partitioned into six parts for motion modeling with masked body parts.}
    \label{fig:body_partition}
\end{figure}

\noindent
\textbf{Motion modeling with masked body parts. } We partition the human body into six parts: upper body, lower body, head, arms, legs, and hips, as shown in~\cref{fig:body_partition}. One part is then randomly chosen to be masked.

This technique enables the pose encoder to learn human kinematics and anatomy more comprehensively than joint-level masking by capturing coordination not only among individual joints but also across multiple body parts. When masked body part motion modeling is combined with multi-modal alignment prediction, the pose encoder further learns the spatial relationships of human motion within the context of corresponding actions. This masking technique also simulates real-world scenarios where parts of the human body may be occluded in a camera's field of view.

\noindent
\textbf{Motion modeling with masked time windows. } In addition to masked body parts, we propose another masking technique that masks the motion in the temporal domain with a varying window size between $[T_1, T_2]$. The starting frame of this masked time window also varies and is randomly chosen.

\begin{figure}
    \centering
    \includegraphics[width=0.55\linewidth]{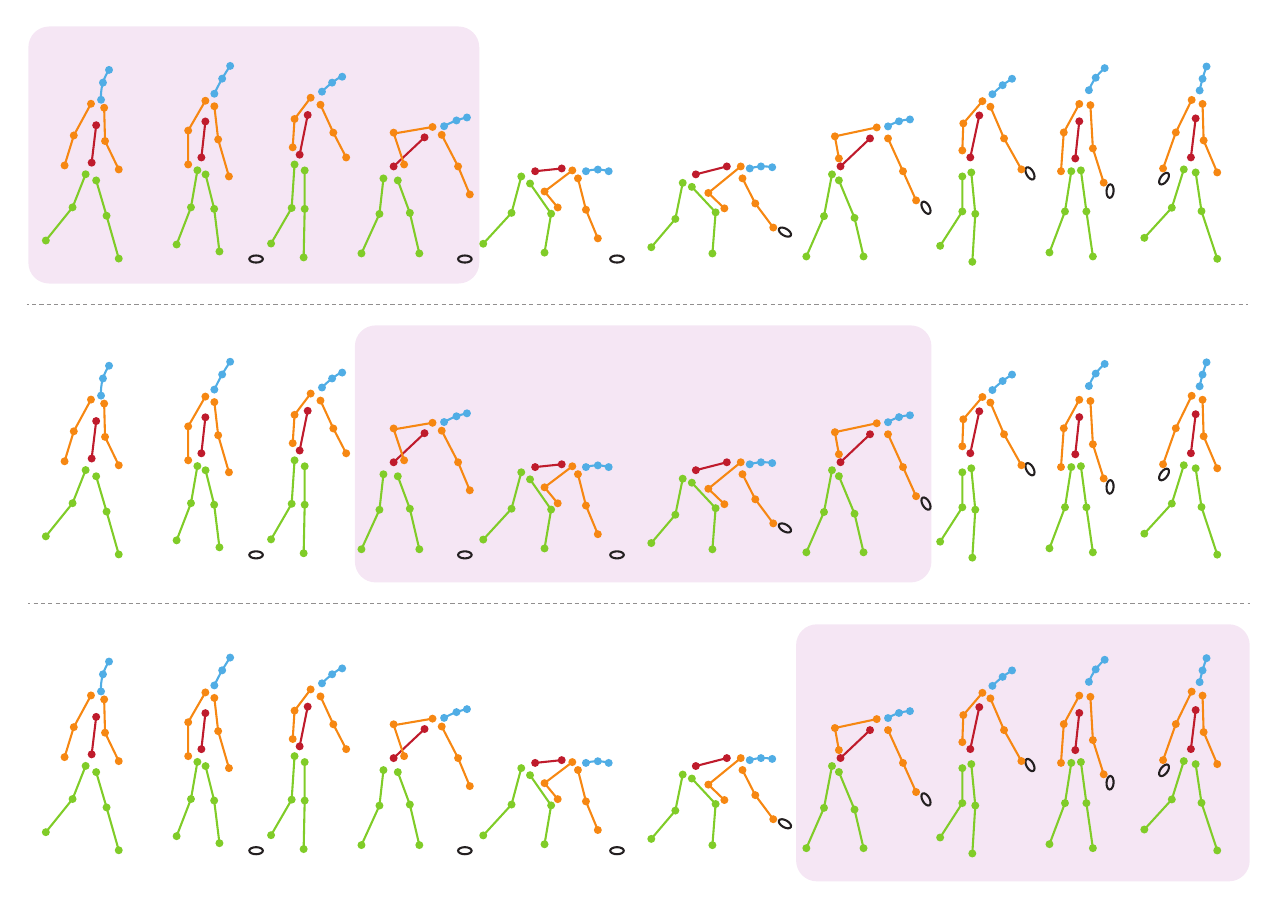}
    \caption{Masking different segments of a human motion has distinct effects. Top: masking the initial part of the motion. Middle: masking the middle segment. Bottom: masking the latter part of the motion.}
    \label{fig:time_masking}
\end{figure}

The method can be interpreted as follows: masking different segments of motion in the time domain serves distinct purposes. Masking the initial part of the motion helps infer the cause of the action. Masking the middle segment allows us to model the transition between different motion states. Masking the latter part aids in predicting the motion’s intention and reconstructing human motion based on that predicted intention.~\Cref{fig:time_masking} provides a visualization of these interpretations.

\noindent
\textbf{Multi-modal alignment prediction. } The dual-stream encoders are optimized together by contrasting the global representation of text embeddings and pose embeddings at sample $j$:

\begin{equation}
   s_{p,w}^j = \frac{exp(h_P^j \cdot h_{W+}^j / \tau)}{\sum_{i=0}^K exp(h_P^j \cdot h_{W_i}^j / \tau)},
\end{equation}
where $h_{W+}^j$ is the corresponding text description of the action, $h_{W_i}^j$ for $i \in [0,1,...,K]$ includes one positive sample $h_{W+}^j$ and $K$ negative samples, and $\tau$ is the temperature.

Since motion-and-language datasets~\cite{BABEL:CVPR:2021, kit} are highly imbalanced and human action recognition tasks are inherently non-deterministic, namely, motion and action labels do not have a one-to-one correspondence, we replace the cross-entropy in the infoNCE loss with a combination of focal loss~\cite{focal_loss} and KL divergence. The modified loss function is as follows:

\begin{equation}
    \mathcal{L}_{con} = \sum_{j=1}^M (1 - s_{p,w}^j)^\gamma y_j log(\frac{y_j}{s_{p,w}^j}),
\end{equation}
where $M$ is the number of overall samples and $y_j$ is the ground-truth similarity score.

\noindent
\textbf{Objective function. } During pretraining, we adopt several masking strategies: joint-level and frame-level masking 50\% of the time to capture spatial consistency among neighboring joints and temporal consistency across adjacent frames; random body part masking 25\% of the time; and random time window masking 25\% of the time. Additionally, we introduce noises into the 2D pose sequences.

Alongside the multi-modal alignment loss, we perform 3D pose reconstruction from the corrupted 2D poses. The discrepancy between the predicted 3D pose sequence $\mathbf{X}$ and the ground-truth 3D pose sequence $\hat{\mathbf{X}}$ is penalized using the following loss functions:

\begin{equation}
    \mathcal{L}_{3D} = \sum_{k=1}^M \sum_{t=1}^T \sum_{j=1}^J ||\mathbf{\hat{X}_{t,j}^k} - \mathbf{X_{t,j}^k}||_2,
\label{eq:3d}
\end{equation}

\begin{equation}
      \mathcal{L}_{v} = \sum_{k=1}^M \sum_{t=1}^T \sum_{j=1}^J ||\mathbf{\hat{V}_{t,j}^k} - \mathbf{V_{t,j}^k}||_2,
\label{eq:velocity}
\end{equation}
Where $\mathbf{\hat{V}_{t}} = \mathbf{\hat{X}_{t}} - \mathbf{\hat{X}_{t-1}}$ and $\mathbf{{V_{t}}} = \mathbf{{X_{t}}} - \mathbf{{X_{t-1}}}$.

The final pretraining loss is computed by combining the multi-modal alignment loss and the reconstruction loss functions as follows:

\begin{equation}
    \mathcal{L} = \lambda_{con} \mathcal{L}_{con} + \lambda_{3D} \mathcal{L}_{3D} + \lambda_v \mathcal{L}_{v} .
\end{equation}

\subsection{Fine-tuning}
During pretraining, the parameters of both the pose encoder and text encoder are updated, but during fine-tuning, only the pose encoder is fine-tuned on the target dataset. Specifically, 2D poses are detected from videos and then lifted to 3D poses. The fine-tuning process is supervised by the reconstruction losses as specified in~\cref{eq:3d} and~\cref{eq:velocity}.

\begin{table*}[ht]
\centering
\caption{Quantitative comparison of 3D human pose estimation on the Human3.6M dataset using detected 2D poses. The per-action performance is shown under P1, and the average metrics across P1, P2, and P3 are summarized. All methods use a temporal window of 243 frames. \dag indicates the use of temporal information, * represents diffusion-based methods, and $\diamond$ indicates pretraining-based methods. Results from other models are directly quoted from the respective papers.}
\resizebox{\textwidth}{!}{%
\begin{tabular}{lccccccccccccccc|ccc}
\toprule
Method & Dir. & Disc. & Eat & Greet & Phone & Photo & Pose & Pur. & Sit & SitD. & Smoke & Wait & WalkD. & Walk & WalkT. &  P1 ↓ & P2 ↓ & P3 ↓\\
\midrule
 P-STMO~\cite{p-stmo} \dag & 38.9 & 42.7 & 40.4 & 41.1 & 45.6 & 49.7 & 40.9 & 39.9 & 55.5 & 59.4 & 44.9 & 42.2 & 42.7 & 29.4 & 29.4 & 42.8 & 34.4 & 54.9
 \\
MixSTE~\cite{mixste} \dag & 37.6 & 40.9 & 37.3 & 39.7 & 42.3 & 49.9 & 40.1 & 39.8 & 51.7 & 55.0 & 42.1 & 39.8 & 41.0 & 27.9 & 27.9 & 40.9 & 32.6 & 52.2 \\
GLA-GCN~\cite{gla-gcn} \dag & 41.3 & 44.3 & 40.8 & 41.8 & 45.9 & 54.1 & 42.1 & 41.5 & 57.8 & 62.9 & 45.0 & 42.8 & 45.9 & 29.4 & 29.9 & 44.4 & 34.8 & 58.3 \\
STCFormer~\cite{STCFormer} \dag & 38.4 & 41.2 & 36.8 & 38.0 & 42.7 & 50.5 & 38.7 & 38.2 & 52.5 & 56.8 & 41.8 & 38.4 & 40.2 & 26.2 & 27.7 & 40.5 & 31.8 & 53.3 \\
KTPFormer~\cite{peng2024ktpformer} \dag & 37.3 & 39.2 & 35.9 & 37.6 & 42.5 & 48.2 & 38.6 & 39.0 & 51.4 & 55.9 & 41.6 & 39.0 & 40.0 & 27.0 & 27.4 & 40.1 & 31.9 & 51.8 \\
TCPFormer~\cite{tcpformer} \dag & -- & -- & -- & -- & -- & -- & -- & -- & -- & -- & -- & -- & -- & -- & -- & 37.9 & 31.7 & -- \\
RePose~\cite{RePose} \dag & 34.6 & 36.8 & 35.0 & \textbf{31.3} & 38.8 & \textbf{44.7} & \textbf{35.3} & 35.5 & 47.5 & 50.9 & 38.4 & \textbf{35.5} & 34.9 & \textbf{23.9} & 25.0 & \textbf{36.5} & \textbf{30.7} & 47.7\\
DiffPose~\cite{diffpose} \dag * & \textbf{33.2} & \textbf{36.6} & \textbf{33.0} & 35.6 & \textbf{37.6} & 45.1 & 35.7 & 35.5 & 46.4 & 49.9 & \textbf{37.3} & 35.6 & 36.5 & 24.4 & \textbf{24.1} & 36.9 & -- & 47.1\\


D3DP~\cite{D3DP} \dag * & 37.3 & 39.5 & 35.6 & 37.8 & 41.3 & 48.2 & 39.1 & 37.6 & 49.9 & 52.8 & 41.2 & 39.2 & 39.4 & 27.2 & 27.1 & 39.5 & 31.6 & 50.3 \\
ActionPrompt~\cite{actionPrompt} & 37.7 & 40.2 & 39.8 & 40.6 & 43.1 & 48.0 & 38.8 & 38.9 & 50.8 & 63.2 & 42.0 & 40.0 & 42.0 & 30.5 & 31.6 & 41.8 & -- & 54.0 \\
MotionBERT~\cite{motionbert2022} \dag $\diamond$ & 36.1 & 37.5 & 35.8 & 32.1 & 40.3 & 46.3 & 36.1 & 35.3 & 46.9 & 53.9 & 39.5 & 36.3 & 35.8 & 25.1 & 25.3 & 37.5 & -- & 49.0 \\

\textbf{LangPose} \dag $\diamond$ & 35.6 & 37.5 & 35.3 & 31.7 & 40.1 & 45.4 & 36.5 & \textbf{33.4} & \textbf{46.2} & \textbf{49.3} & 39.0 & 36.0 & \textbf{34.4} & 24.6 & 25.1 & 36.7 & 31.3 & \textbf{47.0}\\
\bottomrule
\end{tabular}%
}
\label{tab:h36m}
\end{table*}



\begin{table}[htp]
    \centering
    \caption{Quantitative comparison of 3D human pose estimation on the MPI-INF-3DHP dataset. 
    All methods use a temporal window of 81 frames. \dag indicates the use of temporal information, 
    * represents diffusion-based methods, and $\diamond$ indicates pretraining-based methods. 
    Results from other models are directly quoted from the respective papers.}
    \small 
    \setlength{\tabcolsep}{3pt} 
    \renewcommand{\arraystretch}{0.9} 
    \begin{tabular}{lccc}
        \toprule
        Method & PCK $\uparrow$ & AUC $\uparrow$ & P1 $\downarrow$\\
        \midrule
        P-STMO~\cite{p-stmo} \dag           & 97.9 & 75.8 & 32.2 \\
        GLA-GCN~\cite{gla-gcn} \dag         & 98.5 & 79.1 & 27.7 \\
        STCFormer~\cite{STCFormer} \dag     & 98.7 & 83.9 & 23.1 \\
        DiffPose~\cite{diffpose} \dag *     & 98.0 & 75.9 & 29.1 \\
        D3DP~\cite{D3DP} \dag *             & 97.7 & 78.2 & 29.7 \\
        KTPFormer~\cite{peng2024ktpformer} \dag * & 98.9 & 85.9 & 16.7 \\
        MotionAGFormer-L~\cite{MotionAGFormer_2024_WACV} \dag & 98.2 & 85.3 & 16.2 \\
        TCPFormer~\cite{tcpformer} \dag     & \textbf{99.0} & \textbf{87.7} & \textbf{15.0} \\
        RePose~\cite{RePose} \dag & 98.3 & 86.7 & \underline{15.5} \\
        \midrule
        LangPose (Ours) \dag $\diamond$   & \underline{98.9} & \underline{87.0} & \underline{15.5} \\
        \bottomrule
    \end{tabular}
    \label{tab:mpi_inf_3dhp}
\end{table}

\section{Experiments}

\subsection{Pretraining}
The BABEL~\cite{BABEL:CVPR:2021} dataset is a large dataset that includes language labels describing the actions performed in mocap sequences. It provides both sequence-level labels, which describe the overall action in the sequence, and frame-level labels, which detail fine-grained actions in every frame. Each frame in the sequence can have multiple corresponding fine-grained action labels. The motion sequences in the BABEL dataset are sourced from the large mocap dataset AMASS~\cite{AMASS:ICCV:2019}, which uses a common parameterization framework. Instead of obtaining videos and extracting 2D skeleton sequences from the mocap dataset, we project the 3D skeletons extracted from parametric models to 2D from both a side view and a front view, assuming an orthographic camera, as done in~\cite{motionbert2022}. We align the body keypoint definitions with those of Human3.6M and convert the camera coordinates to pixel coordinates using the approach outlined in~\cite{data_pose_normalization}. 

Each positive data pair for pretraining consists of a fine-grained frame-level action label and its corresponding motion sequence. For sequences labeled as ``transition'', we concatenate the motion sequence before and after the ``transition'', and use a template ``transit from A to B'' to construct new action labels. A and B are the corresponding action labels before and after the ``transition''. Negative data pairs are generated by randomly associating action labels with unmatched motion sequences.

The overall pretraining framework consists of three layers of pretrained BERT~\cite{devlin2018bert} as the text encoder and five layers of pretrained MotionBERT~\cite{motionbert2022} as the pose encoder. The network is trained on a single NVIDIA H100 GPU with a batch size of 16 and a sequence length of 243 frames and for 300 epochs using an AdamW optimizer. The loss weights are set to $\lambda_{con}=0.5$, $\lambda_{3D}=1$, and $\lambda_{v}=20$, and the model is trained for 300 epochs. Please refer to the Supplementary Material for more implementation details.

\subsection{Fine-tuning for 3D human pose estimation}

During fine-tuning for 3D human pose estimation, the text encoder is no longer required; only the pose encoder is fine-tuned on the target test dataset.

\noindent
\textbf{Datasets and evaluation metrics.} We evaluated our methods on the public dataset Human3.6M~\cite{h36m_pami} and MPI-INF-3DHP~\cite{mono-3dhp2017}.

Human3.6M~\cite{h36m_pami} contains 3.6 million video frames of human motion performed by professional actors in an indoor environment. Following previous works~\cite{motionbert2022, peng2024ktpformer}, we use subjects 1, 5, 6, 7, and 8 for fine-tuning, and subjects 9 and 11 for testing. We report results using the mean per joint position error (MPJPE), which measures the average Euclidean distance between predicted and ground-truth joint positions, typically in millimeters. We also report the Procrustes-aligned MPJPE (P-MPJPE), where MPJPE is calculated after applying Procrustes alignment to the predicted and ground-truth positions. Futhermore, we observe that three actions are consistently the most challenging to predict across state-of-the-art methods as shown in~\cref{tab:h36m}: \textit{Sitting} (involving external objects), \textit{Sitting Down} (prone to self-occlusions), and \textit{Taking Photo} (less repeatable motion). We also find that these actions are the most heavily occluded, according to a rough estimation of their visibility\footnote{See the Supplementary Material for details of the visibility estimation.}. We denote the average MPJPE over these three actions as H-MPJPE. For simplicity, we refer to MPJPE, P-MPJPE, and H-MPJPE as P1, P2, and P3, respectively.

MPI-INF-3DHP~\cite{mono-3dhp2017} is another large-scale public dataset. This dataset contains recordings from 14 cameras capturing 8 actors performing 8 activities for the training set and 7 activities for evaluation. Following the setting of previous work~\cite{p-stmo, MotionAGFormer_2024_WACV, peng2024ktpformer}, our evaluation metrics included the area under the curve (AUC), percentage of correct keypoints (PCK), and mean per-joint position error (MPJPE).


\noindent
\textbf{Performance comparison on the Human3.6M dataset. } We evaluated LangPose on the Human3.6M dataset~\cite{h36m_pami}, and reported P1, P2, and P3 errors in millimeters. The 2D skeletons are extracted using the Stacked Hourglass network~\cite{stacked_hourglass}, and the pose encoder is fine-tuned on the Human3.6M training set. As shown in~\cref{tab:h36m}, LangPose achieves competitive performance compared to prior temporal models, with particularly large improvements on the three most challenging actions. This highlights the effectiveness of incorporating semantic information when spatial and temporal consistency alone is insufficient. Moreover, LangPose achieves performance comparable to diffusion-based methods, while being more efficient during inference, as diffusion models rely on iterative refinement.

Among all baselines, MotionBERT~\cite{motionbert2022} is most closely related to our framework in terms of its pretraining and fine-tuning strategy. As shown in~\cref{tab:h36m}, our model outperforms MotionBERT by a large margin, especially on the hardest actions, demonstrating the benefits of integrating semantic information during pretraining. It is worth noting that the motion sequences from BABEL~\cite{BABEL:CVPR:2021} originate from the AMASS dataset~\cite{AMASS:ICCV:2019}, which is already included in the pretraining data of MotionBERT. Although our pretraining includes additional action labels to facilitate semantic alignment, no extra motion sequences are used beyond what is available to MotionBERT.

\noindent
\textbf{Performance comparison on the MPI-INF-3DHP dataset. } To evaluate the generalization capability of LangPose, we assessed its performance on the MPI-INF-3DHP dataset. Compared to Human3.6M, the test set of MPI-INF-3DHP features more diverse motion types and greater variation in camera viewpoints. For instance, MPI-INF-3DHP includes sports-related poses that are absent from Human3.6M. Additionally, approximately 67\% of the poses in Human3.6M fall under the ``stand/pose'' category, while only 25\% correspond to ``sit'' and 8\% to ``crouch''---the latter being more complex and challenging to model~\cite{mono-3dhp2017}.

Following prior works~\cite{MotionAGFormer_2024_WACV, peng2024ktpformer}, we fine-tuned the pose encoder using ground-truth 2D poses as input. As shown in~\cref{tab:mpi_inf_3dhp}, LangPose achieves SOTA level performance, obtaining a PCK of 98.9\%, an AUC of 87.0\%, and an MPJPE of 15.5~mm. These results surpass most existing methods, including both temporal-based and diffusion-based approaches, highlighting the strong generalization ability of Pose.L

\begin{table}[ht]
\centering
\caption{Ablation study of LangPose on MPI-INF-3DHP and Human3.6M datasets. Results are reported using P1 and P3 metrics (in mm). We refer to masked motion modeling as MMM and multi-modal alignment prediction as MAP.}
\resizebox{\columnwidth}{!}{%
\begin{tabular}{lccccc}
\toprule
\multirow{2}{*}{Configuration} & \multicolumn{2}{c}{Modules enabled} & MPI (P1) & H36M (P1) & H36M (P3) \\
\cmidrule(lr){2-3}
 & MMM & MAP &  &  &  \\
\midrule
Baseline              & \xmark & \xmark & 16.7 & 37.5 & 49.0 \\
Baseline + MMM        & \cmark & \xmark & 16.2 & 36.8 & 47.6 \\
Baseline + MAP        & \xmark & \cmark & 15.7 & 37.3 & 47.9 \\
MMM + MAP (ours)      & \cmark & \cmark & \textbf{15.5} & \textbf{36.7} & \textbf{47.0} \\
\bottomrule
\end{tabular}
}
\label{tab:ablation_masking_with_map}
\end{table}

\begin{table}[ht]
  \centering
  \caption{Ablation study comparing the proposed modified InfoNCE loss with the original InfoNCE formulation on MPI-INF-3DHP (abbrev. MPI) and Human3.6M (abbrev. H36M). MMM denotes masked motion modeling. Results are reported using P1 and P3 metrics (mm).}
  \label{tab:loss_compare}
  \begin{adjustbox}{max width=\columnwidth}
  \begin{tabular}{p{0.42\columnwidth} >{\centering}p{0.18\columnwidth} >{\centering}p{0.18\columnwidth} >{\centering\arraybackslash}p{0.18\columnwidth}}
    \toprule
    Method (Loss) & MPI (P1) & H36M (P1) & H36M (P3) \\
    \midrule
    Baseline (w/o InfoNCE) & 16.7 & 37.5 & 49.0 \\
    Baseline + original InfoNCE + MMM & 16.3 & 37.5 & 48.5 \\
    Baseline + MMM + \textbf{Modified InfoNCE (ours)} & \textbf{15.5} & \textbf{36.7} & \textbf{47.0} \\
    \bottomrule
  \end{tabular}
  \end{adjustbox}
\end{table}

\noindent
\textbf{Ablation study. } To verify the effectiveness of our proposed framework, we conducted ablation studies on the MPI-INF-3DHP dataset ($T$ = 81) using ground-truth 2D poses as inputs and on the Human3.6M dataset ($T$ = 243) using detected 2D poses.~\Cref{tab:ablation_masking_with_map} presents the results for each component added to our framework, reported as P1 for the Human3.6M dataset and as both P1 and P3 for the MPI-INF-3DHP dataset (in mm). 

For the baseline, we used the pretrained MotionBERT model fine-tuned on the MPI-INF-3DHP and Human3.6M datasets, respectively. The pretrained MotionBERT weights were obtained from the official GitHub page. On MPI-INF-3DHP, pretraining with the proposed masked time windows and masked body parts motion modeling (MMM) reduced the error by 0.5mm compared to the baseline. Incorporating random frame and joint-level masking, random noise, and multi-modal alignment prediction further reduced the error by 1.0mm (MAP). Combining these strategies (MMM + MAP) resulted in a total error reduction of 1.2mm from the baseline. Similarly, on Human3.6M, pretraining with masked time windows and masked body parts motion modeling reduced the error by 0.7mm, and adding random frame and joint-level masking, noise, and multi-modal alignment reduced the error by an additional 0.2mm. The combination of these techniques led to a total error reduction of 0.8mm from the baseline. P3 is additionally reported to highlight the performance of the proposed modules on the most difficult actions in Human3.6M, where the joint contribution of masked motion modeling and semantic alignment proves especially beneficial.

To further validate the effectiveness of our proposed loss design, we conduct an additional ablation study by comparing the modified InfoNCE loss with the original InfoNCE formulation. Specifically, in the modified InfoNCE loss, the cross-entropy term is replaced with a KL-divergence objective, and a focal loss component is incorporated to better handle hard examples. As shown in ~\cref{tab:loss_compare}, this modification leads to consistent performance improvements across both benchmarks. Moreover, we observe that the original InfoNCE loss tends to suffer from overfitting during training, whereas our modified InfoNCE loss successfully mitigates this issue, providing a stronger optimization signal and yielding more robust representations for downstream 3D human pose estimation.


\subsection{Qualitative results on global representation learning}

\begin{figure}[h]
    \centering 
    \begin{subfigure}{0.23\textwidth}
        \includegraphics[width=1\linewidth]{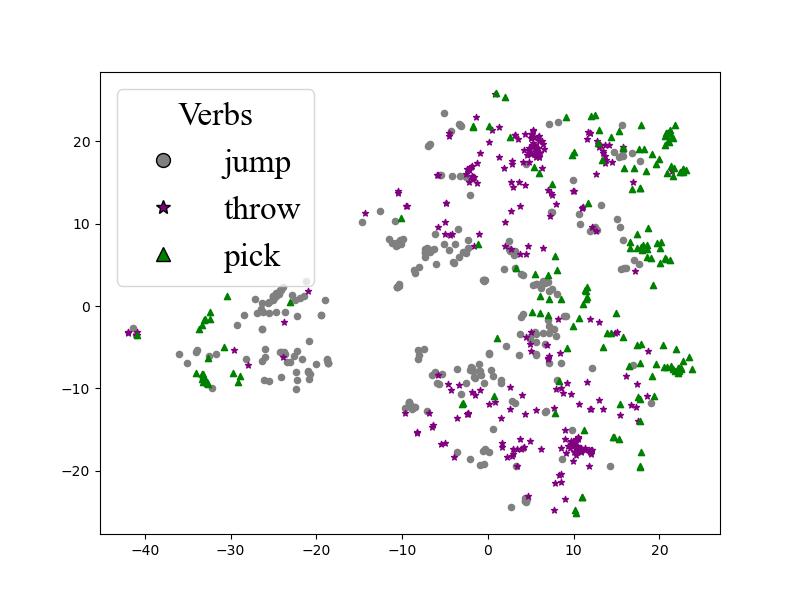}
        \label{fig:1}
    \end{subfigure}\hfil 
\begin{subfigure}{0.23\textwidth}
  \includegraphics[width=1\linewidth]{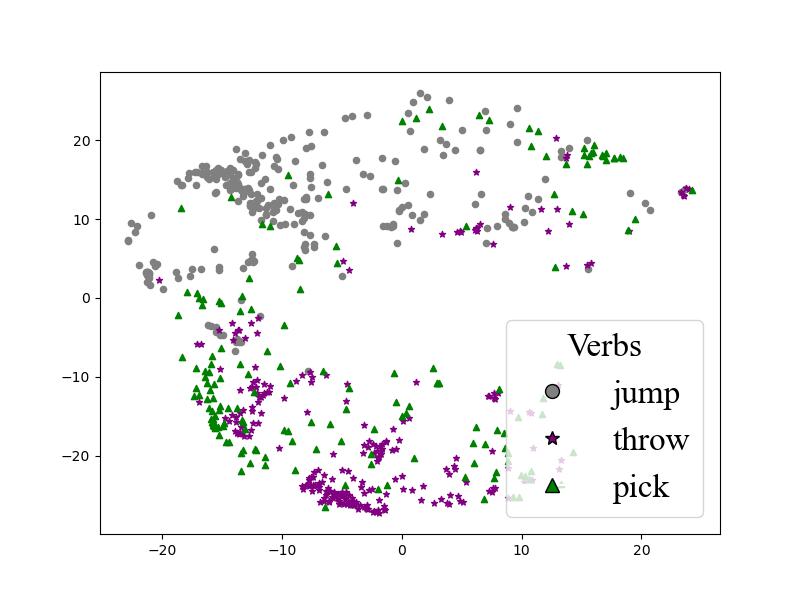}
  \label{fig:2}
\end{subfigure}
\medskip
\begin{subfigure}{0.23\textwidth}
  \includegraphics[width=1\linewidth]{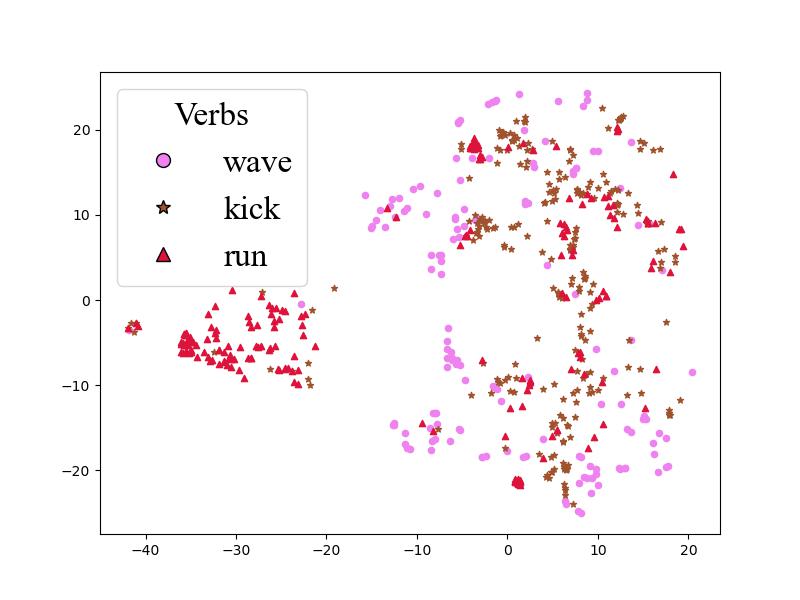}
  \label{fig:3}
\end{subfigure}
\hfil
\begin{subfigure}{0.23\textwidth}
  \includegraphics[width=1\linewidth]{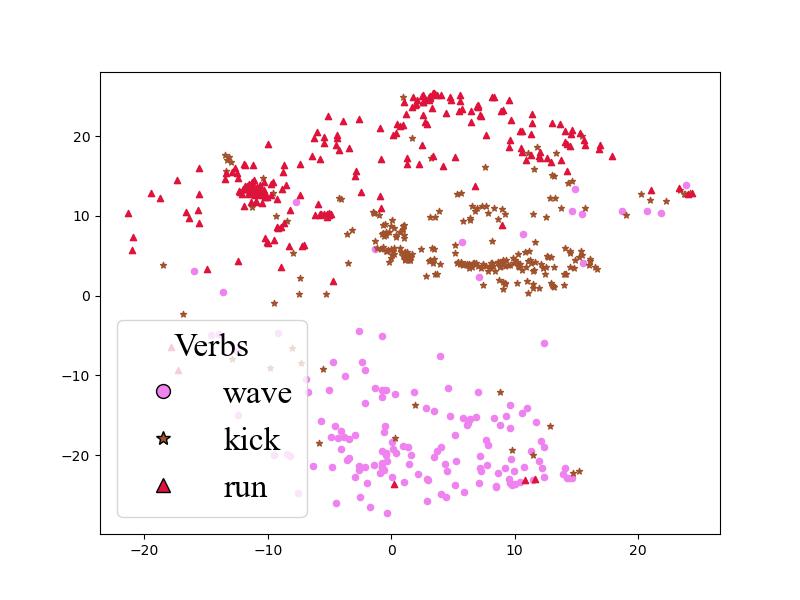}
  \label{fig:4}
\end{subfigure}
\medskip
\begin{subfigure}{0.23\textwidth}
  \includegraphics[width=1\linewidth]{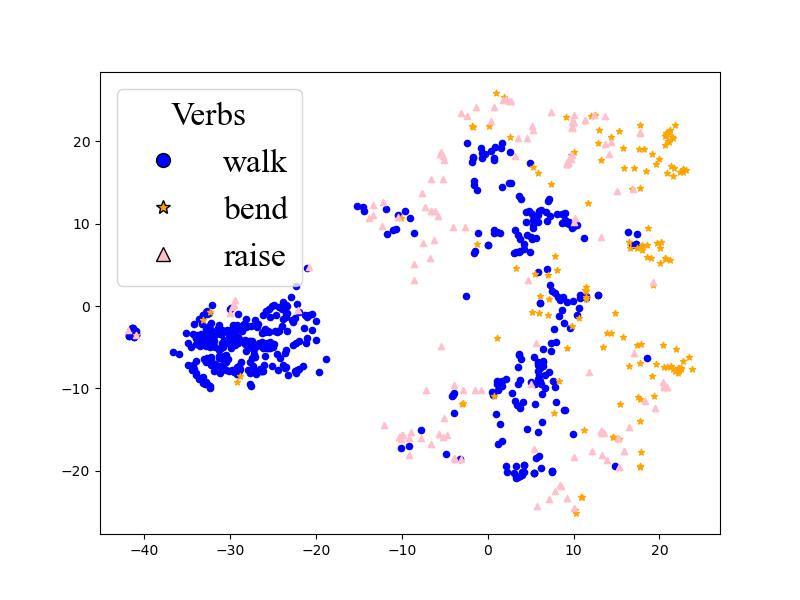}
  \caption{MotionBERT}
  \label{fig:5}
\end{subfigure}\hfil 
\begin{subfigure}{0.23\textwidth}
  \includegraphics[width=1\linewidth]{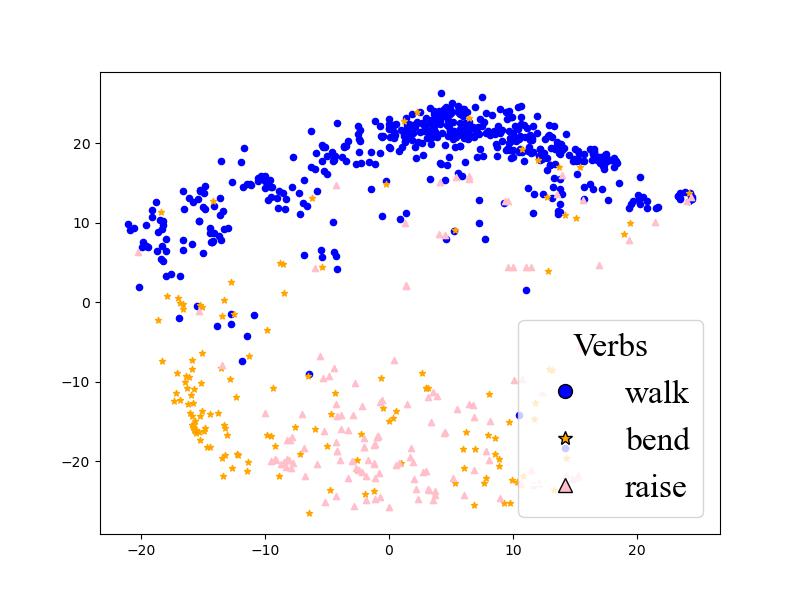}
  \caption{LangPose}
  \label{fig:6}
\end{subfigure}
\caption{The t-SNE visualization of global representations of motion embeddings on the BABEL dataset: on the left (a) are embeddings obtained through random joint-level and frame-level masking, as proposed by MotionBERT~\cite{motionbert2022}, with average pooling of all pose features in the sequence; on the right (b) are embeddings obtained through LangPose.}
\label{fig:vis_global}
\end{figure}

To verify LangPose's ability to learn rich semantic information about actions, we visualize the global representations of pose embeddings in~\cref{fig:vis_global}. Ideally, semantically similar actions should have closer embeddings, while different actions should be farther apart. For instance, `throw' and `pick' both emphasize upper body movements, and thus their representations are close. In constrast, `jump', which emphasizes lower body movement, is distinctly separate from them. Similarly, `run' and `kick' are similar from leg movements and are positioned near each other, while `wave' focused on the upper body, is more distant. Additionally, `bend' and `raise' are closely related, whereas `walk' is farther from these actions. 

As shown in~\cref{fig:vis_global}, LangPose effectively groups semantically similar actions closer together while pushing semantically different actions farther apart. In contrast, MotionBERT's global representation lacks a clear pattern, with significant overlap between both similar and distinct actions. Please note that there is some diversity within the same action due to variations in concurrent actions and different forms and styles of the actions. For example, ``kicking while jumping" differs from ``kicking while standing," and the single action label with the verb `kick' include motion sequences for both ``kicking a soccer ball" or ``kicking with the right leg back and up".

\section{Discussion}

Learning sufficient action semantics during text embed-
ding alignment with motion representation requires a large
dataset. When the action classes in the pretraining dataset
differ from those in the target dataset, the benefits of fus-
ing action knowledge through pretraining may diminish\footnote{See the Supplementary Material for more discussion.}. Despite this limitation, beyond 3D human pose estimation, LangPose can be extended to other human motion-related tasks, such as motion prediction and completion. For motion prediction, the model can use historical motion to predict intentions, thereby guiding future movements. Similarly, for motion completion, the model can learn transitions between starting and ending poses, constrained by the corresponding actions. The framework's ability to learn rich semantic information about actions also has potential benefits for downstream tasks such as motion captioning from noisy inputs and generation from text descriptions. These applications are reserved for future work.
\section{Conclusion}

In this work, we demonstrate how understanding human motion from an action perspective can enhance 3D human pose estimation. We propose a novel pretraining and fine-tuning strategy that embeds action cues during the pretraining stage. Experimental results confirm the effectiveness of the proposed masked motion modeling and multi-modal alignment tasks. Additionally, the ability to incorporate semantic information into global representation learning shows potential for benefiting downstream tasks such as human motion captioning, generation, and prediction.

\begin{figure}[htp] 
  \centering
  \begin{subfigure}{0.3\linewidth}
    \centering
    \includegraphics[width=1\linewidth]{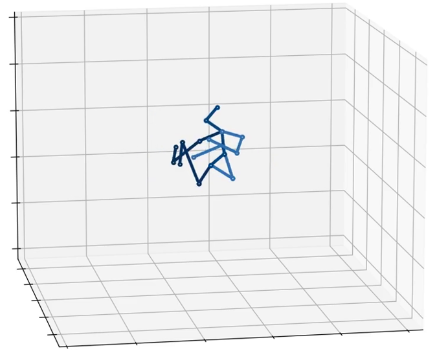} 
    \label{fig:subfig1}
  \end{subfigure}
  \hfil
  \begin{subfigure}{0.3\linewidth}
    \centering
    \includegraphics[width=1\linewidth]{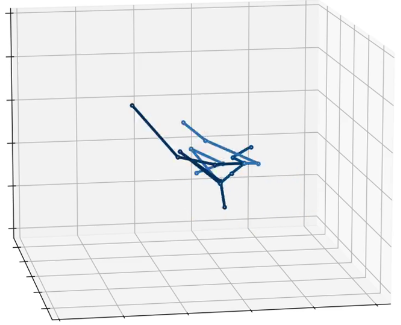} 
    \label{fig:subfig2}
  \end{subfigure}
    \hfil
  \begin{subfigure}{0.3\linewidth}
    \centering
    \includegraphics[width=0.95\linewidth]{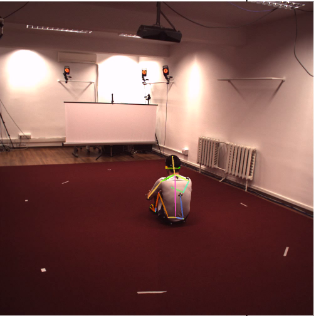} 
    \label{fig:subfig2}
  \end{subfigure}

\medskip
\begin{subfigure}{0.3\linewidth}
    \centering
    \includegraphics[width=1\linewidth]{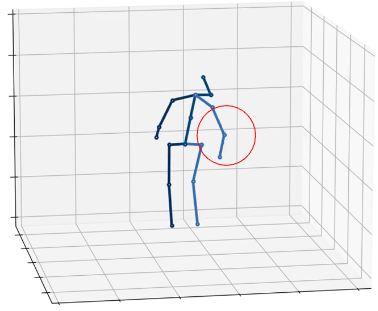} 
    \label{fig:subfig1}
  \end{subfigure}
  \hfil
  \begin{subfigure}{0.3\linewidth}
    \centering
    \includegraphics[width=1\linewidth]{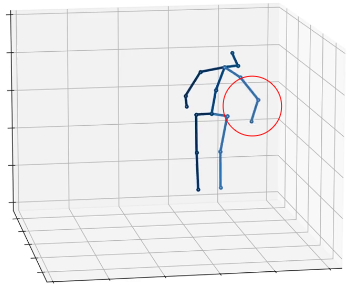} 
    \label{fig:subfig2}
  \end{subfigure}
    \hfil
  \begin{subfigure}{0.3\linewidth}
    \centering
    \includegraphics[width=0.95\linewidth]{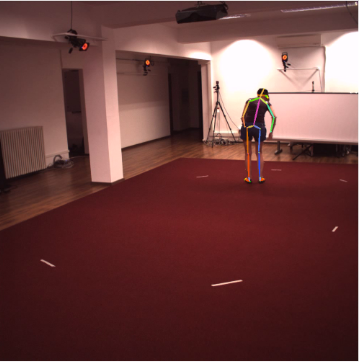} 
    \label{fig:subfig2}
  \end{subfigure}

  \medskip
\begin{subfigure}{0.3\linewidth}
    \centering
    \includegraphics[width=1\linewidth]{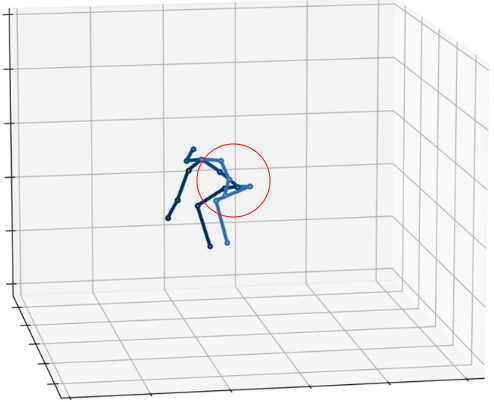} 
    \label{fig:subfig1}
  \end{subfigure}
  \hfil
  \begin{subfigure}{0.3\linewidth}
    \centering
    \includegraphics[width=1\linewidth]{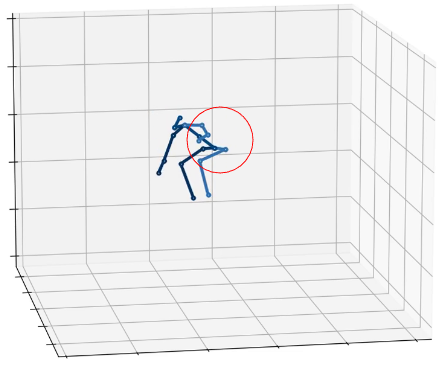} 
    \label{fig:subfig2}
  \end{subfigure}
    \hfil
  \begin{subfigure}{0.3\linewidth}
    \centering
    \includegraphics[width=0.95\linewidth]{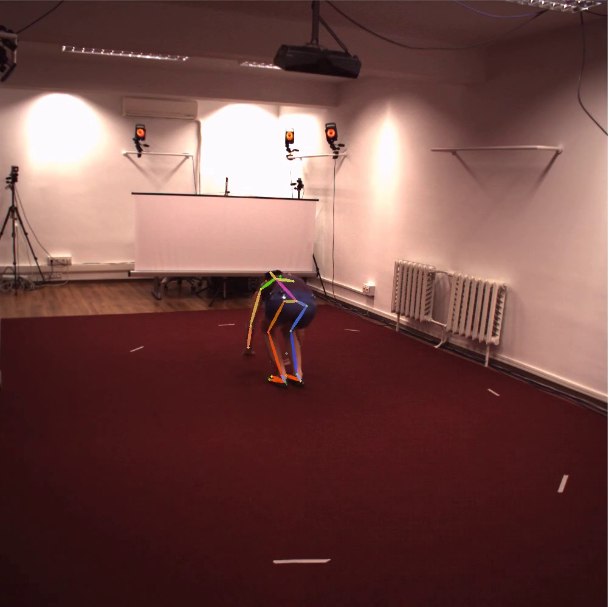} 
    \label{fig:subfig2} 
  \end{subfigure}
  
  \caption{Qualitative results of LangPose compared to the baseline model MotionBERT. The left column shows the results from LangPose, the middle column shows the results from MotionBERT for the same frames, and the right column displays the corresponding original 2D pose inputs.}
  \label{fig:qualitative_compare}
\end{figure}
{
    \small
    \bibliographystyle{ieeenat_fullname}
    \bibliography{main}
}

\end{document}